\PassOptionsToPackage{table}{xcolor}
\documentclass[sigconf]{acmart}

\copyrightyear{2026}
\acmYear{2026}
\setcopyright{cc}
\setcctype{by}
\acmConference[MM '26] {Proceedings of the 34th ACM International Conference on Multimedia}{November 10--14, 2026}{Rio de Janeiro, Brazil.}
\acmBooktitle{Proceedings of the 34th ACM International Conference on Multimedia (MM '26), November 10--14, 2026, Rio de Janeiro, Brazil}
\acmISBN{979-8-4007-2213-4/2026/11}
\acmDOI{10.1145/3767308.3836551}
\usepackage{multirow} 
\usepackage{enumitem}
\begin{document}

%%
%% The "title" command has an optional parameter,
%% allowing the author to define a "short title" to be used in page headers.
\title{SuppreSensing: Expert-Guided Feature Recalibration and Discrepancy Augmentation for Multimodal Object Detection}

%%
%% The "author" command and its associated commands are used to define
%% the authors and their affiliations.
%% Of note is the shared affiliation of the first two authors, and the
%% "authornote" and "authornotemark" commands
%% used to denote shared contribution to the research.
% \author{Ben Trovato}
% \authornote{Both authors contributed equally to this research.}
% \email{trovato@corporation.com}
% \orcid{1234-5678-9012}
% \author{G.K.M. Tobin}
% \correspondingauthor
% \authornotemark[1]
% \email{webmaster@marysville-ohio.com}
% \affiliation{%
%   \institution{Institute for Clarity in Documentation}
%   \city{Dublin}
%   \state{Ohio}
%   \country{USA}
% }

% \author{Lars Th{\o}rv{\"a}ld}
% \affiliation{%
%   \institution{The Th{\o}rv{\"a}ld Group}
%   \city{Hekla}
%   \country{Iceland}}
% \email{larst@affiliation.org}

% \author{Huifen Chan}
% \affiliation{%
%   \institution{Beijing University of Posts and Telecommunications}
%   \city{Haidian Qu}
%   \state{Beijing Shi}
%   \country{China}}

% \author{Julius P. Kumquat}
% \correspondingauthor
% \affiliation{%
%   \institution{Beijing University of Posts and Telecommunications}
%   \city{New York}
%   \country{USA}}
% \email{jpkumquat@consortium.net}

\author{Xin Wu}
\affiliation{%
  \institution{Beijing University of Posts and Telecommunications}
  \city{Beijing}
  \country{China}}
\email{xin.wu@bupt.edu.cn}
\orcid{0000-0002-1733-3560}
  
\author{Zhenyu Gao}
\affiliation{%
  \institution{Beijing University of Posts and Telecommunications}
  \city{Beijing}
  \country{China}}
\email{zhenyugao@bupt.edu.cn}
\orcid{0009-0006-8504-5612}
  
\author{Qiankun Zhang}
\affiliation{%
  \institution{Beijing University of Posts and Telecommunications}
  \city{Beijing}
  \country{China}}
\email{1004221119@email.cugb.edu.cn}
\orcid{0009-0005-0851-4289}
  
\author{Shaoyong Guo}
% \correspondingauthor
\affiliation{%
  \institution{Beijing University of Posts and Telecommunications}
  \city{Beijing}
  \country{China}}
\email{syguo@bupt.edu.cn}
\orcid{0000-0003-2033-8431}

\begin{abstract}
Multimodal object detection in remote sensing faces challenges due to semantic heterogeneity and modality-specific noise interference. To this end, we propose SuppreSensing, which reformulates multimodal fusion as a selective collaboration process that jointly models shared information and modality-specific cues. SuppreSensing first designs an Expert-driven Multimodal Feature Recalibration (EMFR) module, which reformulates shared-consensus extraction as an input-adaptive multi-expert selection process to alleviate the symmetry trap in multimodal fusion. Complementing this, a modality-specific attribute augmentation strategy is employed to enhance specific modality features by modeling bidirectional discrepancy patterns, mitigating cross-modal heterogeneity. Furthermore, we propose an Expert-driven Customized Feature Purification (ECFP) module based on a "specialized inspection-comprehensive analysis-diagnostic update" physical examination paradigm to iteratively filter redundancies and reinforce task-relevant semantics. Extensive experiments on the DroneVehicle and VEDAI datasets demonstrate that SuppreSensing achieves state-of-the-art detection performance. Cross-domain evaluations on natural scene datasets (FLIR and LLVIP) further validate its superior robustness and generalization capability across diverse environmental conditions. The code will be available at \url{https://github.com/Victoria-xin1009/SuppreSensing} for reproducibility.   
\end{abstract}

%%
%% The code below is generated by the tool at http://dl.acm.org/ccs.cfm.
%% Please copy and paste the code instead of the example below.
%%

%%
%% Keywords. The author(s) should pick words that accurately describe
%% the work being presented. Separate the keywords with commas.

\maketitle

\section{Introduction}
Multimodal object detection in remote sensing (RS) is essential for maintaining reliable perception across diverse illumination and degraded environmental conditions. Unlike ground-level snapshots, unmanned aerial vehicle (UAV)-captured RS imagery exhibits large-scale variations and complex backgrounds, further exposing the limitations of unimodal sensors. Visible sensors provide high spatial resolution and rich texture details, but their performance deteriorates in low-light conditions or atmospheric disturbances such as haze \cite{wu2024collaboration}. In contrast, infrared sensors are more robust to such interference and can effectively highlight thermal targets, yet they suffer from low resolution and insufficient fine-grained semantic information. This inherent complementarity motivates visible-infrared fusion for learning robust feature representations. However, semantic discrepancy across modalities and modality-specific noise in RS imagery pose challenges to all-weather object detection.

Although feature-level fusion is now the dominant paradigm, most existing methods\cite{lin2025butter,ding2025theoretical,10.1145/3664647.3680971} still rely on a consistency alignment assumption. Whether through projection-based shared-space alignment \cite{zhao2025freefusion} or attention-based interactions \cite{10.1145/3581783.3612135}, these methods prioritize semantically consistent information across modalities to maintain semantic coherence \cite{wu2025dhanet}. However, this “symmetry trap” tends to suppress discriminative modality-specific cues by treating them as unreliable interference \cite{xiu2025caprecover}. For instance, subtle thermal signatures in infrared imagery or fine-grained texture details in visible data are often weakened or even eliminated during excessive semantic homogenization. As a result, such strong-alignment paradigms struggle to balance the extraction of shared information with the preservation of modality-specific evidence\cite{gao2025sparse}.

\begin{figure}[!t]
  \centering
  \includegraphics[width=\linewidth]{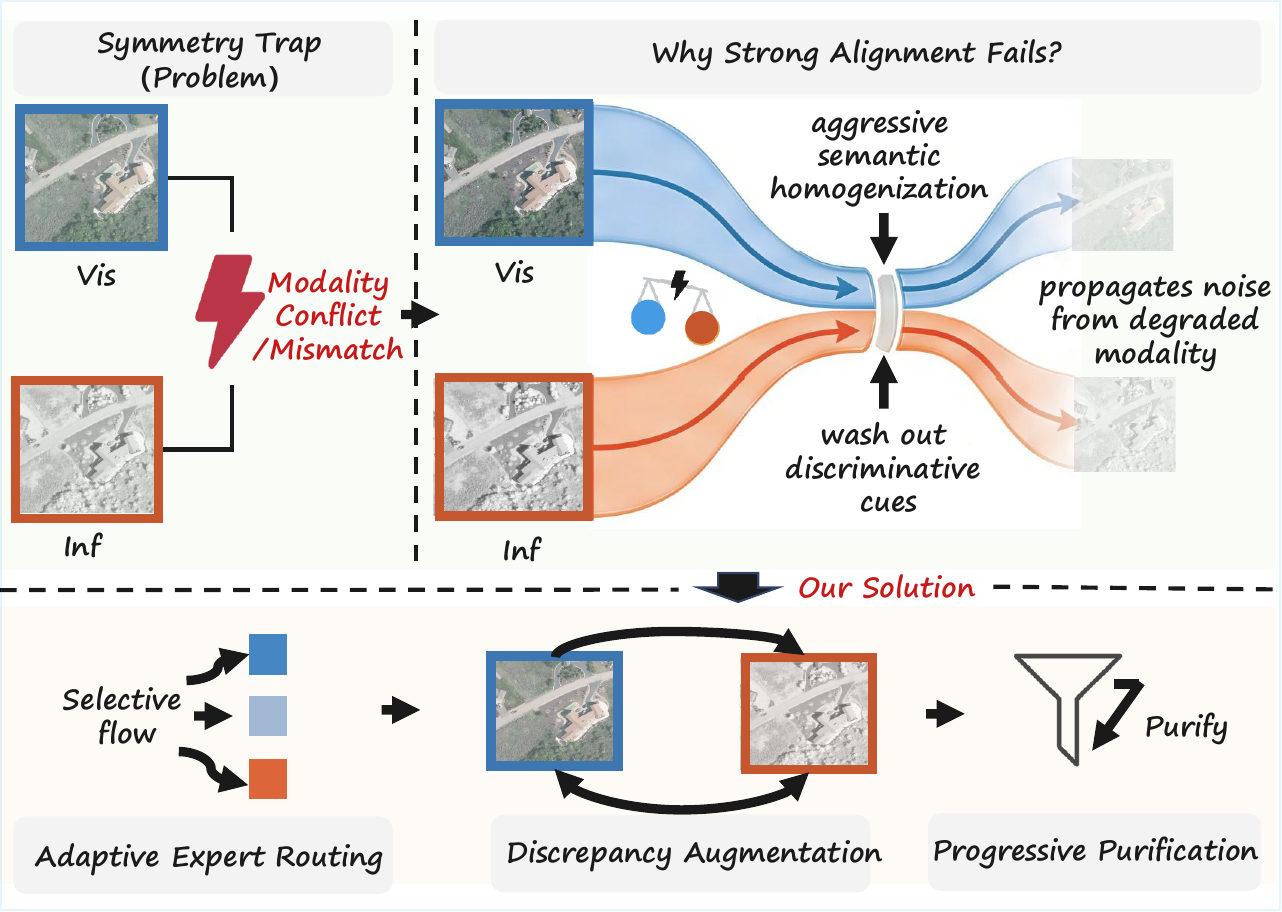}
  \caption{Conceptual comparison between the conventional strong alignment paradigm and our proposed SuppreSensing. Conventional methods are prone to the ``symmetry trap", where semantic homogenization suppresses modality-specific discriminative features (top). In contrast, our method selectively routes features and retains informative modality-specific discrepancies for robust multimodal fusion (bottom).}
  % \Description{A woman and a girl in white dresses sit in an open car.}
  \label{fig1}
\end{figure}

These observations suggest that effective multimodal detection requires selective cross-modal collaboration to capture both shared information and modality-specific evidence. Although MoE \cite{li2025unimoe,park2025resilient}provides a promising framework for dynamic selection, existing methods still struggle to escape the symmetry trap. Most existing methods\cite{wei2025improving,zamfir2025complexity} inherit balanced routing strategies from natural language processing, whereas multimodal imagery exhibits asymmetry under modality-specific degradation. In such cases, strict alignment suppresses modality-specific cues and propagates noise from degraded modalities.

To break this trap, we propose SuppreSensing, which integrates expert-driven feature recalibration and progressive feature purification to selectively exploit shared and modality-specific information, as illustrated in Fig.~\ref{fig1}. Specifically, we first propose expert-driven multimodal feature recalibration (EMFR), which transforms consensus extraction into an input-adaptive selection process to suppress noise while preserving fine-grained modality-specific differences. To further reduce heterogeneity across modalities, we employ a modality-specific attribute augmentation strategy that explicitly models bidirectional discrepancy patterns to reinforce distinctive modality-specific cues. Finally, we design an expert-driven customized feature purification (ECFP) module based on a "specialized inspection-comprehensive analysis-diagnostic update" paradigm to progressively filter redundant information and strengthen task-relevant semantics. 

The main contributions of this work are as follows:

\begingroup
\setlength{\listisep}{2pt}   % 必须放在 \begin{itemize} 之前
\setlength{\partopsep}{0pt}
\setlength{\itemsep}{1pt}
\setlength{\parsep}{0pt}
\setlength{\parskip}{0pt}
\begin{itemize}[
    leftmargin=*,
    itemindent=0pt,
    listparindent=0pt,
    labelsep=0.4em,
    topsep=2pt,
    itemsep=1pt,
    parsep=0pt,
    partopsep=0pt
]
    \item We propose SuppreSensing, a novel expert-guided method for multimodal object detection in RS imagery. It reformulates multimodal fusion as a selective collaboration process that jointly models shared information and modality-specific cues.
    \item We propose an Expert-driven Multimodal Feature Recalibration (EMFR) module, which reformulates shared-consensus extraction as an input-adaptive multi-expert selection process. EMFR alleviates the symmetry trap in multimodal MoE by suppressing modality-specific noise while preserving informative modality-specific discrepancies.
    \item We propose an Expert-driven Customized Feature Purification (ECFP) module based on a ``specialized inspection-comprehensive analysis-diagnostic update". Through iterative expert-guided refinement, ECFP progressively removes redundant information and reinforces task-relevant semantics, enabling continuous purification of multimodal representations.
    \item Extensive experiments on DroneVehicle, VEDAI, LLVIP, and FLIR demonstrate that SuppreSensing achieves state-of-the-art performance and strong robustness under diverse illumination and environmental conditions.
\end{itemize}
\endgroup

\section{Related Work}
\subsection{Multimodal Object Detection}
Existing multimodal object detection methods in RS mainly differ in how and where information from different modalities is fused. Among them, pixel-level fusion \cite{wang2025real} and decision-level fusion \cite{chen2022multimodal} are representative fusion paradigms. Pixel-level fusion preserves low-level physical cues through early-stage integration, but it is vulnerable to spatial misalignment and noise propagation. Decision-level fusion, by contrast, maintains independent modality-specific inference branches and aggregates only final predictions, which improves tolerance to modality inconsistency but weakens the modeling of cross-modal semantic interactions.

To bridge this gap, feature-level fusion has become the dominant paradigm, offering a more favorable trade-off between representation capacity and robustness to interference. Recent studies have focused on designing more effective multimodal interaction mechanisms, such as attention-based feature calibration (e.g., C2former \cite{yuan2024c2former}) and spatial-semantic enhancement strategies (e.g., DEYOLO (Dual-feature-enhancement YOLO) \cite{chen2024deyolo}), to improve the quality of fused representations. However, they are still largely grounded in a strong-alignment paradigm. Such methods force heterogeneous modalities into a unified semantic space and assume that different modalities should contribute roughly symmetrically during fusion. Although strong alignment may enhance semantic consistency, it can also suppress modality-specific cues that are crucial for detection, such as subtle thermal responses in infrared imagery and fine-grained texture patterns in optical data. Some studies \cite{zhao2025removal,he2023multispectral} attempt to alleviate this issue through coarse-to-fine feature selection or conflict-aware learning, but they still fail to reconcile effective multimodal interaction with the preservation of modality-specific attributes.
\begin{figure*}
    \centering
    \includegraphics[width=7in]{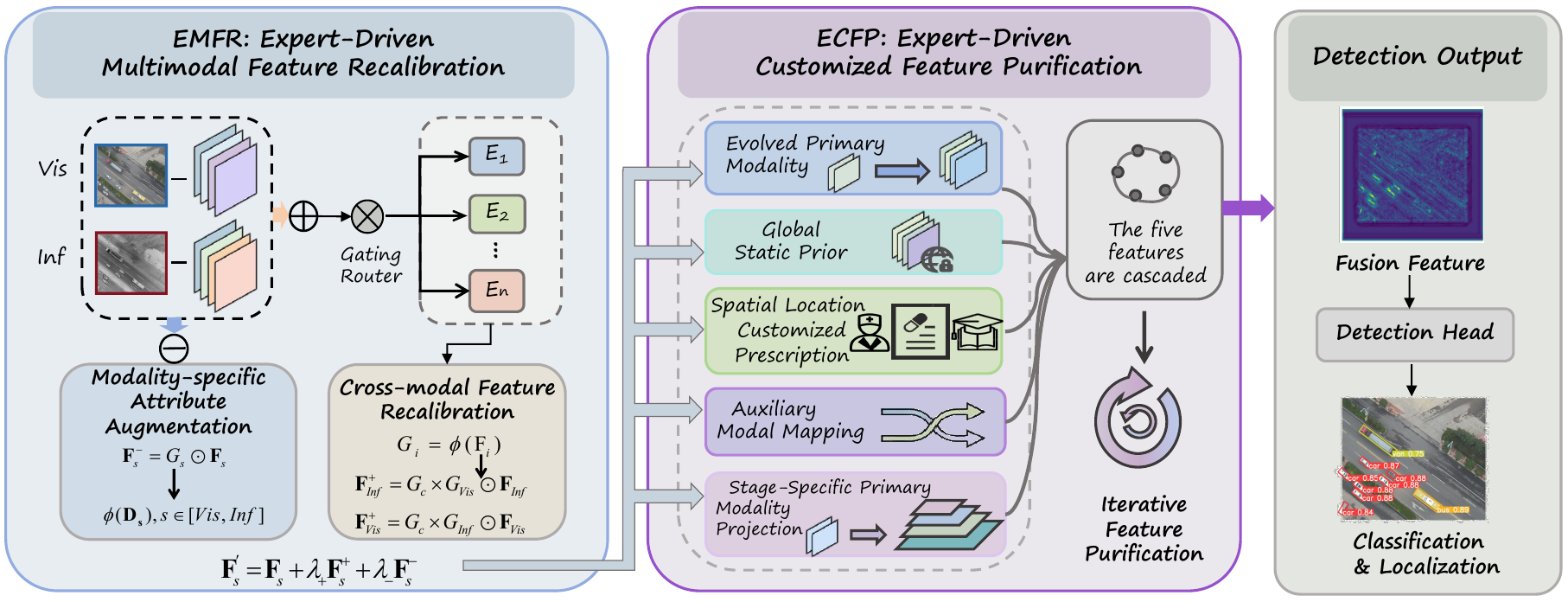}%   
    \caption{Pipeline of the SuppreSensing method. The EMFR module first processes the visible and infrared features for expert-driven recalibration, which extracts shared consensus and enhances modality discrepancies. Subsequently, ECFP iteratively purifies the recalibrated features through five customized paths before final classification and localization.}
    \label{fig:outline}
\end{figure*}

\subsection{Mixture of Experts Mechanism }
MoE is composed of multiple expert sub-networks and a gating router, which together enable an input-dependent sparse activation mechanism for dynamic computation allocation. Owing to this design, MoE offers scalability and adaptive modeling ability for complex heterogeneous data. The idea of MoE was first introduced by Jacobs et al. \cite{jacobs1991adaptive} in 1991. After a relatively dormant period, it regained broad attention in natural language processing with the sparsely gated MoE layer proposed by Shazeer et al. \cite{shazeer2017outrageously}.

Compared to traditional static fusion paradigms, MoE facilitates dynamic expert activation and weight assignment based on input modal features and environmental contexts, enabling more specialized feature extraction and adaptive cross-modal synthesis. Cao et al. \cite{cao2023multi} enhanced global context awareness and local granularity by dynamically aggregating multi-receptive-field features. Liu et al. \cite{11071289} employed dynamic gating for expert allocation, along with frequency-decoupled fusion modules, to achieve environment-adaptive detection. However, real-world multimodal data often exhibit significant discrepancies in sensing mechanisms, signal-to-noise ratios, spatial resolutions, and environmental sensitivity. Moreover, the task-specific utility of each modality varies significantly across different scenarios. MoE-SPDF \cite{dai2026moe} developed a scene-aware dynamic MoE fusion framework for infrared-visible joint representation under extreme illumination. Feng et al. \cite{feng2025text} further leveraged dynamic routing to strengthen the complementarity between visible texture details and infrared thermal cues. Despite these advances, most existing multimodal MoE methods still adopt a symmetric routing strategy inherited from language models, which can introduce noise from low-quality modalities and thus undermine the discriminability and robustness of the fused representation.

\section{Proposed Method}

As illustrated in Fig.~\ref{fig:outline}, SuppreSensing comprises EMFR for multimodal feature recalibration and ECFP for customized feature purification.

\subsection{Expert-Driven Multimodal Feature Recalibration (EMFR)}
\begin{figure}[!t]
  \centering
  \includegraphics[width=\linewidth]{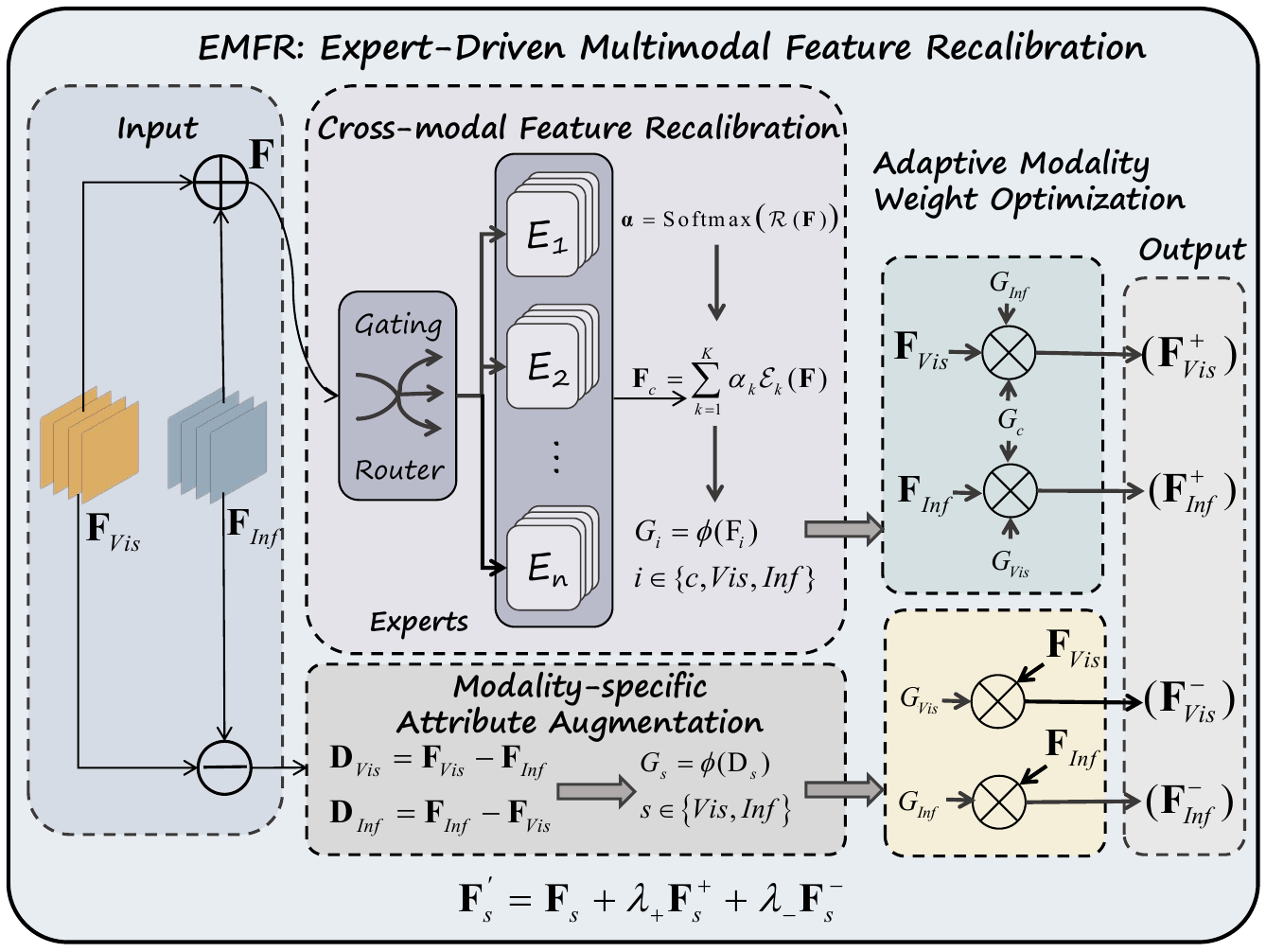}
  \caption{Illustration of the EMFR module. It decomposes multimodal data into two complementary pathways. One extracts shared consensus through dynamic expert routing, while the other models bidirectional discrepancies. The recalibrated features are then aggregated to produce refined features for downstream spatial purification. }
  % \Description{A woman and a girl in white dresses sit in an open car.}
  \label{fig:EMFR}
\end{figure}

For object detection in multimodal RS images, modality conflict frequently arises. For example, visible images offer rich features but suffer from localized overexposure under intense illumination, whereas infrared images avoid overexposure but exhibit lower resolution. As illustrated in Fig.~\ref{fig:EMFR}, EMFR performs expert-driven dynamic channel recalibration for unimodal feature enhancement through two components: cross-modal feature recalibration and modality-specific attribute augmentation.

\subsubsection{Cross-modal Feature Recalibration (CFR)}
We employ a multimodal cue-driven gating router to dynamically activate specialized experts in the channel subspace. This design transforms shared consensus extraction from a fixed mapping into an input-adaptive expert selection process. It allows the model to adjust the extraction strategy based on scene context. Under strong daylight, the model may rely more on infrared cues to compensate for visual overexposure. Under low-light conditions, it may place more emphasis on the thermal saliency of Inf imagery. Based on the inferred scene tendency, the router selects appropriate experts to mine modality-shared information. The resulting consensus is then injected into the visual and infrared branches through cross-gating. In this way, MoE improves the quality of shared consensus extraction and provides a more reliable basis for subsequent cross-modal gated injection. 

Given multimodal features $ \mathbf{F}_{s} \in {\mathbb{R}^{C \times H \times W}}, s \in {Vis, Inf}$, where $C$, $H$ and $W$ denote the channel number, height and width of feature maps, we first form an integrated representation $ \mathbf{F} =  \mathbf{F}_{Vis} +  \mathbf{F}_{Inf}$. The router predicts expert weights as $
\boldsymbol{\alpha} = \mathrm{Softmax}\big(\mathcal{R}( \mathbf{F})\big)$, where $\mathcal{R}(\cdot )$ is a lightweight router network. 

The final shared consensus $\mathbf{F}_c$ is obtained by computing an adaptive weighted sum of the outputs from the $K$ homogeneous experts $\mathcal{E}_k(\cdot)$:
% The shared consensus is obtained by 

\begin{equation}
\mathbf{F}_c = \sum_{k=1}^{K} \alpha_k \mathcal{E}_k(\mathbf{F}),
\end{equation}
where $K$ is the number of experts, $\mathcal{E}_k$ is the $k$-th expert, and $\alpha_k$ is the activation weight of $\mathcal{E}_k$. 

By utilizing a channel-wise attention operator $\phi(\cdot)$, the resulting consensus is injected into the unimodal and multimodal branches to yield the gating weights $G_{i} = \phi(F_i)$ for $i \in \{c, Vis, Inf\}$. Subsequently, the cross-modal feature recalibration is

\begin{equation}
\mathbf{F}^{+}_{Inf} = G_{c} \times G_{Vis} \odot  \mathbf{F}_{Inf} \qquad  \mathbf{F}^{+}_{Vis} =G_{c} \times G_{Inf} \odot  \mathbf{F}_{Vis},
\end{equation}
where $\odot$ is the element-wise multiplication.

\subsubsection{Modality-specific Attribute Augmentation (MAA)}
To address the heterogeneity across modalities, the specific modality features are further enhanced. We compute bidirectional discrepancy features as
\begin{equation}
\mathbf{D}_{Vis} = { \mathbf{F}}_{Vis} - \mathbf{F}_{Inf}, \qquad
\mathbf{D}_{Inf} = { \mathbf{F}}_{Inf} - \mathbf{F}_{Vis}.
\end{equation}

The discrepancy features are further refined by a nonlinear transformation to preserve informative discrepancy patterns while suppressing irrelevant disturbances. The refined feature $\mathbf{F}^{-}_{s} $ is 

\begin{equation}
\mathbf{F}_s^- = G_s \odot \mathbf{F}_s = \phi(\mathbf{D}_s) \odot \mathbf{F}_s.
\end{equation}

Finally, the recalibrated multimodal feature is

\begin{equation}
 \mathbf{F}_{s}^{\prime} =  \mathbf{F}_{s} + \lambda_{+}  \mathbf{F}_{s}^{+} + \lambda_{-}  \mathbf{F}_{s}^{-},
\end{equation}
where $\lambda_{+}$ and $\lambda_{-}$ are the learnable parameters.

\subsection{Expert-Driven Customized Feature Purification (ECFP)} 

\begin{figure}[!t]
  \centering
  \includegraphics[width=\linewidth]{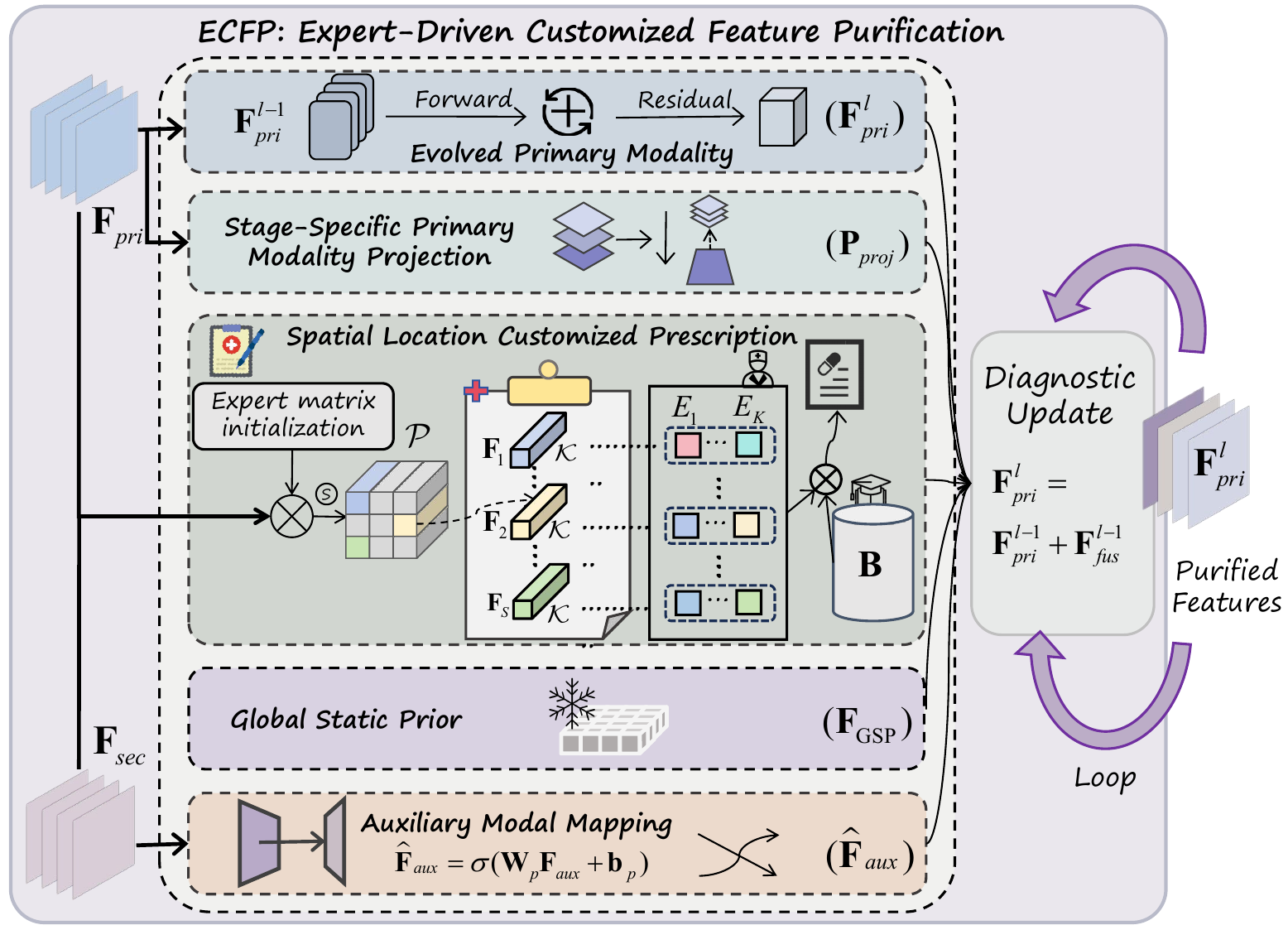}
  \caption{Illustration of the ECFP module. It integrates five customized feature pathways to perform spatially adaptive refinement across heterogeneous regions. Through iterative updates, the module progressively improves fusion features and reduces background interference. }
  % \Description{A woman and a girl in white dresses sit in an open car.}
  \label{fig:ECFP}
\end{figure}

Inspired by physical examination, ECFP follows a ``specialized inspection-comprehensive analysis-diagnostic update'' paradigm, as illustrated in Fig.~\ref{fig:ECFP}. It transforms multimodal features into five customized pathways serving as diagnostic indicators for each spatial location. The evolved primary modality preserves the infrared features from the previous layer as an evolutionary baseline, while the global static prior calibrates cross-modal distribution shifts. The spatial customized prescription generates location-specific expert responses. Meanwhile, auxiliary modal mapping and stage-specific modality projection align the auxiliary features with the primary modality and adapt the primary features across stages to maintain semantic consistency, respectively. These components are concatenated into a comprehensive report, which residually updates the primary features. Through iterative updates, ECFP suppresses redundant information and strengthens task-relevant semantics.

\subsubsection{Five-Path Customized Feature}
Given the modality-specific dependencies, multimodal remote sensing interpretation should determine the primary and auxiliary modalities based on specific imaging conditions. In this path, we design five feature branches to achieve customized processing of the primary and auxiliary modal features.

\textbf{Spatial Location Customized Prescription (SLCP):} RS images are characterized by complex backgrounds and feature variations across different spatial locations. This path provides a set of customized prescriptions for each spatial location to assist in calibrating these spatial-dimensional discrepancies. Specifically, this path introduces a MoE mechanism to design a feature-learning paradigm called ``joint consultation and customized prescription". By feeding the learned multimodal features into $K$ expert subspaces, this path diagnoses features at each spatial location and formulates a customized prescription.

\begin{table*}[!t]
    \centering
    \caption{Statistical comparison of the four evaluated multimodal datasets.}
    % \resizebox{\textwidth}{!}{
     \resizebox{.95\linewidth}{!}{
    \begin{tabular}{c|c|c|c|c|c|c|c}
    \toprule[0.8pt]
    \multicolumn{2}{c|}{Dataset} & Image Pairs & Resolution & Classes & Annotation format & Alignment & Lighting Condition\\
    \hline 
    \multirow{2}{*}{Remote Sensing}&DroneVehicle & 28439 & 840×712 & 5 & OBB & No & Day, night, and dark night \\
    &VEDAI & 1210 & 1024×1024 & 8 & HBB & Yes & Day \\ \hline 
    \multirow{2}{*}{Natural Scene}&LLVIP & 15488 & 1280×1024 & 1 & HBB & Yes & Night, dark night  \\
    &FLIR & 5142 & 640×512 & 3 & HBB & Yes & Day, night \\
    \bottomrule[0.8pt]
\end{tabular}
 }
\label{tab:four}
\end{table*}

The process is divided into two stages: symptom diagnosis and customized prescription.

\textbf{1) Spatial Feature Symptom Diagnosis}

Given the primary and auxiliary modality features $\mathbf{F}_{pri}, \mathbf{F}_{sec} \in \mathbb{R}^{B \times S \times C}$, where $B$ denotes the batch size,$S = H \times W$ denotes the number of spatial locations. The router network jointly projects two modality features to generate pixel-wise ``symptom" representation $\mathbf{G}_{rout}$:

\begin{equation}
\mathbf{G}_{rout} = Concat(\mathbf{W}_x(\mathbf{F}_{pri}), \mathbf{W}_y(\mathbf{F}_{sec})) \in \mathbb{R}^{B \times S \times 2C}, 
\end{equation}
where $\mathbf{W}_x$ and $\mathbf{W}_y$ are router projection layers containing non-linear activations.

Subsequently, a frozen global and orthogonal expert matrix $\mathbf{E} \in \mathbb{R}^{2C \times K}$ is defined as the expertise signature for the $K$ experts. By computing the inner product between the symptom representation $\mathbf{G}_{rout}$ at each spatial location and each expert $K$, we evaluate the match between the feature demands and each expert's specific expertise. The final matching weights $\mathbf{M} \in \mathbb{R}^{B \times S \times K}$ are calculated as follows:

\begin{equation}
\mathbf{M} = \text{Softmax}\left(\frac{\mathbf{G}_{rout} E}{\tau} + \epsilon\right),
\end{equation}
where $\tau$ is the temperature parameter, and $\epsilon$ is a noise term. Here, $\mathbf{M}$ determines how an expert's influence is applied to the pathological features at location $s$. This process effectively realizes the expert ''joint consultation."

\textbf{2) Spatial Feature Customized Prescription}

Once the spatial consultation weights are determined, we extract the corresponding expertise for each expert. We define the learnable expert knowledge base as $\mathcal{P} \in \mathbb{R}^{K \times \mathcal{K}}$, where $\mathcal{K}$ represents each expert's knowledge dimension. This matrix $\mathbf{\mathcal{P}}$ serves as shared knowledge, acting as the core medical compendium. The model expands the global compendium $\mathcal{P}$ along the spatial dimension $S$. This constructs a candidate feature space $\hat{\mathcal{P}} \in \mathbb{R}^{K \times S \times \mathcal{K}}$. Guided by the matching weights $\mathbf{M}$, the network personalizes the expert knowledge. Ultimately, it formulates a customized feature prescription:

\begin{equation}
\bf{F}_{B,S,\mathcal{K}} = \sum\limits_\mathcal{K} {{{\bf{M}}_{B,S,K}}}  \cdot {{\bf{\hat{\mathcal{P}}}}_{K,S,\mathcal{K}}}.
\end{equation}

\textbf{Auxiliary Modal Mapping (AMM):} Direct fusion of heterogeneous modalities leads to semantic misalignment. To address this, this path projects the auxiliary features $\mathbf{F}_{aux}$ into a latent space through a series of linear and nonlinear transformations. This process calibrates the representational deviation between the auxiliary and primary modalities, ultimately boosting the multi-modal fusion capability. The calibrated feature $\hat{\mathbf{F}}_{aux}$ is

\begin{equation}
\hat{\mathbf{F}}_{aux} = \sigma(\mathbf{W}_p \mathbf{F}_{aux} + \mathbf{b}_p),
\end{equation}
where $\mathbf{W}_p$ and $\mathbf{b}_p$ are learnable linear parameters. The function $\sigma(\cdot)$ represents the non-linear activation. 
 
\textbf{Stage-specific Primary Modality Projection (SPMP):} As the network deepens, primary modality features evolve from low-level texture to high-level semantic representations. The path employs nonlinear projection layers to extract features specific to each stage's depth. This process calibrates semantic consistency across hierarchical levels. The stage-specific projection at $l$ layer is $\mathbf{P}_{proj}^{(l)}$ 

\begin{equation}
\mathbf{P}_{proj}^{(l)} = \begin{cases} 
\Phi_0(\mathbf{F}_{pri}), & l \in \{0, ..., k\} \\ 
\Phi_1(\mathbf{F}_{pri}), & l \in \{k+1, ..., N\} ,
\end{cases}
\end{equation}
\begin{equation}
\Phi_m(\mathbf{F_{pri}}) = \text{Dropout}(\sigma(\mathbf{W}_m \mathbf{F_{pri}} + \mathbf{b}_m)),  m \in {0,1},
\end{equation}
where $\mathbf{W}_m$ and $\mathbf{b}_m$ denote the learnable weights and biases of the linear transformation, while $\sigma(\cdot)$ and $\text{Dropout}(\cdot)$ represent the non-linear activation function and regularization operation, respectively.

\begin{table*}[!htbp]
  \centering
  \caption{Detection accuracy of different methods on VEDAI data. Best results in each column are in bold.}
  \tiny
  \label{tab:vehda}
  % 跨双栏表格使用 \linewidth 或 \textwidth 都可以
  \resizebox{.9\linewidth}{!}{
  \begin{tabular}{@{}lccccccccr@{}}
    \toprule[0.7pt]
    Method & car & pickup & camping & truck & other & tractor & boat & van & mAP@0.5 \\
    \midrule
    FFCA-YOLO\cite{zhang2024ffca}  & 0.896 & 0.857 & 0.787 & 0.857 & 0.486 & 0.818 & 0.615 & 0.670 & 0.748 \\
    SuperYOLO\cite{zhang2023superyolo}  & 0.911 & 0.857 & 0.793 & 0.702 & 0.573 & 0.804 & 0.602 & 0.765 & 0.751 \\
    MMFDet\cite{zhao2025mmfdet}    & 0.883 & 0.785 & 0.816 & 0.598 & 0.635 & \textbf{0.862} & 0.760 & 0.883 & 0.779 \\
    DHANet\cite{wu2025dhanet}     & 0.866 & 0.903 & 0.813 & 0.796 & \textbf{0.791} & 0.609 & 0.848 & 0.676 & 0.782 \\
    AFFNet\cite{chen2025alignment}   & 0.909 & \textbf{0.905} & 0.888 & 0.905 & 0.631 & 0.677 & 0.782 & 0.739 & 0.806 \\
    ADMPF\cite{liu2025aerial}     & \textbf{0.945} & 0.903 & 0.847 & 0.744 & 0.572 & 0.747 & \textbf{0.855} & \textbf{0.919} & 0.816 \\
  
    \rowcolor{gray!20} Ours(SuppreSensing)          & 0.890 & 0.840 & \textbf{0.900} & \textbf{0.933} & 0.727 & 0.802 & 0.762 & 0.866 & \textbf{0.840} \\
    \bottomrule[0.7pt]
  \end{tabular}
  }
\end{table*}

\textbf{Global Static Prior (GSP):} Features undergo drastic distribution shifts across different layers. This path provides each layer with inherent statistical characteristics that are independent of the input content, aiding calibration of cross-modal distribution deviations across layers.

\textbf{Evolved Primary Modality (EPM):} Serving as the core benchmark for the fusion process, the features in this path are derived from the diagnostic update of the previous layer, providing an evolutionary reference with iterative continuity. The current-layer infrared features are updated as :

\begin{equation}
\mathbf{F}_{pri}^{(l)} = \mathbf{F}_{pri}^{(l - 1)} + \mathcal{F}_{fus}(Concat(\mathbf{F}_{EPM}^{(l - 1)}, \mathbf{F}_{GSP}^{(l - 1)}, \mathbf{F}_{B,S,\mathcal{K}}^{(l - 1)}, \hat{\mathbf{F}}_{aux}^{(l - 1)}, \mathbf{P}_{proj}^{(l - 1)})).
\end{equation}

\textbf{Finally}, the five decoupled feature streams are concatenated and aggregated along the channel dimension to form a comprehensive report. % The comprehensive report updates the primary modality features as residuals, completing the diagnostic update. 

\subsubsection{Iterative Feature Purification}
To further filter out redundant information between heterogeneous modalities and enhance the discriminability of shared semantics, this module employs a cyclic iterative purification strategy. Through hierarchical evolution, the components of each stream progressively suppress task-irrelevant noise and yield more discriminative primary-modality representations after each iteration.

\section{Experiments}
\subsection{Experimental Setup}

\noindent\textbf{\textit{Datasets.}} We evaluate our method on four standard benchmarks across two diverse domains to validate its robustness and generalization. For aerial remote sensing scenarios, we utilize DroneVehicle \footnote{\url{https://github.com/VisDrone/DroneVehicle}} (28,439 Vis-Inf pairs, 5 classes, featuring large-scale variations, complex urban scenarios, and varying day/night illumination) and VEDAI(Vehicle Detection in Aerial Imagery) \footnote{\url{https://downloads.greyc.fr/vedai/}} (1210 Vis-Inf pairs, 8 classes, fine-grained small object detection against complex backgrounds with objects occupying merely 0.7\% of pixels). To validate generalization in ground-level natural scenes, we employ the aligned LLVIP \footnote{\url{https://github.com/bupt-ai-cz/LLVIP}} (15,488 Vis-Inf pairs, focusing on pedestrian detection under low-light conditions) and the FLIR dataset \footnote{\url{https://www.flir.com/oem/adas/adas-dataset-form/} }(5,142 Vis-Inf pairs, 3 classes, covering complex day/night autonomous-driving scenarios). These datasets cover diverse sensing modalities, domains, and object characteristics, enabling a comprehensive evaluation of multimodal detection. Detailed statistics are summarized in Table~\ref{tab:four}.

\noindent\textbf{\textit{Evaluation Metrics.}} Following standard object detection protocols, we measure detection accuracy using mean Average Precision at an IoU threshold of 0.5 (mAP@0.5) and averaged over multiple thresholds from 0.5 to 0.95 (mAP@0.5:0.95). Furthermore, we evaluate computational complexity and model size using GFLOPs and parameter count (Params), respectively.

\noindent\textbf{\textit{Implementation Details.}} Our method is implemented in PyTorch and trained on dual NVIDIA RTX 4090 GPUs. The model is trained for 200 epochs with a batch size of 10. We optimize the network using Stochastic Gradient Descent (SGD) with an initial learning rate of 0.01, momentum of 0.937, and weight decay of $5 \times 10^{-4}$.

% \paragraph{\textbf{Implementation Details.}}

% The proposed improved YOLOv8s model was implemented using the PyTorch 2.5.1 framework. All experiments were conducted through parallel training on a server equipped with two NVIDIA GeForce RTX 4090 (24GB) GPUs, running Ubuntu 22.04.5 LTS with CUDA 12.1 support. During the training process, the model was trained for 200 epochs with a batch size of 10. The Stochastic Gradient Descent (SGD) optimizer was employed for parameter optimization, with an initial learning rate of 0.01, a momentum of 0.937, and a weight decay coefficient of 0.0005.

\subsection{Comparison with SOTA Methods}
We evaluate our method against state-of-the-art (SOTA) multimodal object detection methods on the VEDAI (Table~\ref{tab:vehda}) and DroneVehicle (Table~\ref{tab:drone}) benchmarks. Overall, SuppreSensing achieves superior detection performance achieves the best overall mAP on both benchmarks, validating its robustness in suppressing modality-specific noise and bridging semantic heterogeneity.

\begin{table}[!htbp]
  \centering
  \caption{Detection accuracy of different methods on DroneVehicle data. Best results in each column are in bold.}
  \label{tab:drone}
  % 跨栏时稍微收缩一点留白，用 0.85\linewidth 也很合适
  \resizebox{1.0\linewidth}{!}{
  \begin{tabular}{@{}lcccccc@{}}
    \toprule
    Method  & car & truck & bus & van & freight & mAP@0.5 \\
    \midrule
    AFFNet\cite{chen2025alignment}   & 0.902 & 0.660 & 0.881 & 0.488 & 0.509 & 0.688 \\
    DPAL\cite{liu2025dual}    & 0.953 & 0.748 & 0.942 & 0.516 & 0.587 & 0.749 \\
    DHANet\cite{wu2025dhanet}  & 0.958 & 0.769 & 0.913 & 0.593 & 0.643 & 0.775 \\
    CCLDet\cite{shang2025ccldet}  & \textbf{0.977} & 0.754 &\textbf{0.957} & 0.595 & 0.688 & 0.794 \\
    CMIFDF\cite{zhao2025cmifdf} & 0.916 & 0.784 & \textbf{0.957} & 0.668 & 0.648 & 0.795 \\
    NOC-YOLO\cite{zhang2025noc}   & 0.958 & 0.781 & 0.942 & 0.602 & 0.667 & 0.795 \\
    MGMF\cite{wang2024mask}  & 0.914 & 0.701 & 0.911 & \textbf{0.694} & \textbf{0.785} & 0.803 \\

    \rowcolor{gray!20} Ours(SuppreSensing)           & 0.965 & \textbf{0.823} & 0.926 & 0.649 & 0.700 & \textbf{0.812} \\
    \bottomrule
  \end{tabular}
  }
\end{table}

\begin{table}[!htbp] 
  \renewcommand\arraystretch{1.2}
  \centering
  \tiny
  \caption{Detection accuracy of different methods on LLVIP and FLIR natural-scene datasets. Best results are in bold.}
  \label{tab:nal}
  \resizebox{.85\columnwidth}{!}{
  \begin{tabular}{c|c|c}
  \toprule[0.7pt]
    \multicolumn{3}{c}{\textit{\textbf{LLVIP Dataset}}} \\\hline 
    \textbf{Modality} & \textbf{Method} & \textbf{mAP@0.5} \\
    \hline  
    \multirow{7}{*}{Multimodal} 
    & DM-Fusion\cite{xu2024dmfusion} & 0.881 \\
    & DIVFusion\cite{tang2023divfusion} & 0.898 \\
    & MoE-Fusion\cite{cao2023multi} & 0.910  \\
    & CSSA\cite{cao2023multimodal} & 0.943 \\
    & YOLO-Adaptor\cite{fu2024yoloadaptor} & 0.965 \\
    & DHANet \cite{wu2025dhanet} & \textbf{0.977} \\
    \rowcolor{gray!20}&  Ours (SuppreSensing) & \textbf{0.977}\\ 
    \bottomrule[0.7pt]
    \multicolumn{3}{c}{\textit{\textbf{FLIR Dataset}}} \\\hline 
    \textbf{Modality} & \textbf{Method} & \textbf{mAP@0.5}\\
    \hline 
    \multirow{8}{*}{Multimodal} 
    & MMTOD-CG\cite{devaguptapu2019borrow} & 0.614 \\
    & MMTOD-UNIT\cite{devaguptapu2019borrow} & 0.615 \\
    & GAFF\cite{zhang2021guided} & 0.729 \\
    & CFR\cite{zhang2020multispectral} & 0.724 \\
    & BU-ATT\cite{kieu2021bottom} & 0.731 \\
    & BU-LTT \cite{kieu2021bottom} & 0.732 \\
    & DHANet \cite{wu2025dhanet} & 0.743 \\
    \rowcolor{gray!20}&  Ours (SuppreSensing)  & \textbf{0.760} \\ 
    \bottomrule[0.7pt]
  \end{tabular}
  } % 如果上面启用了 \resizebox，这里也要取消注释
\end{table}

\begin{figure*}
    \centering
    \includegraphics[width=5.8in]{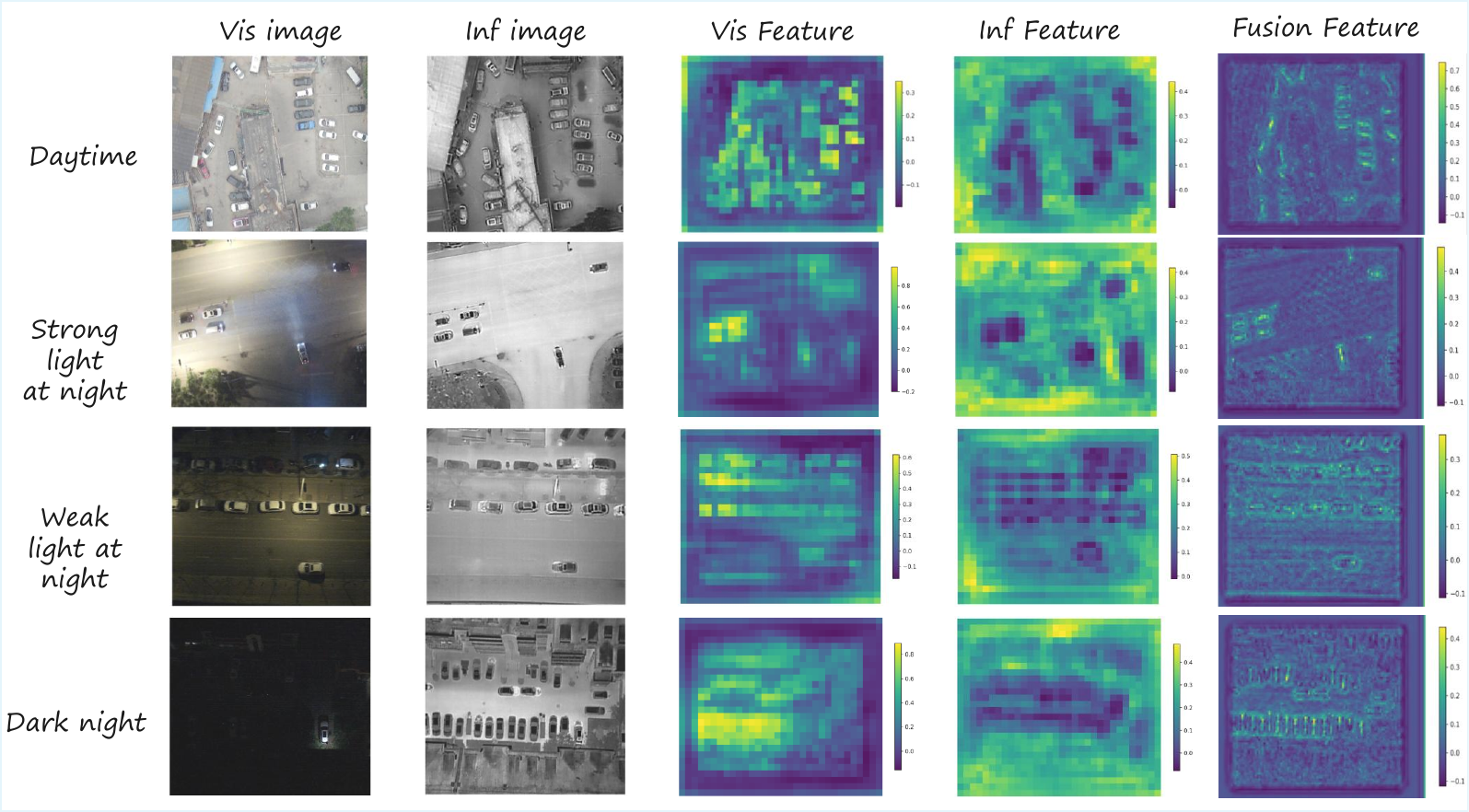}%   
    \caption{Feature map visualizations under varying illumination conditions in DroneVehicle data. The first four columns show the original images and single-modality features, while the last column shows that the fused features suppress environmental distractors and maintain stable object responses. }
     \label{fig:M-AAM}
\end{figure*}

\begin{figure}[!t]
  \centering
  \includegraphics[width=\linewidth]{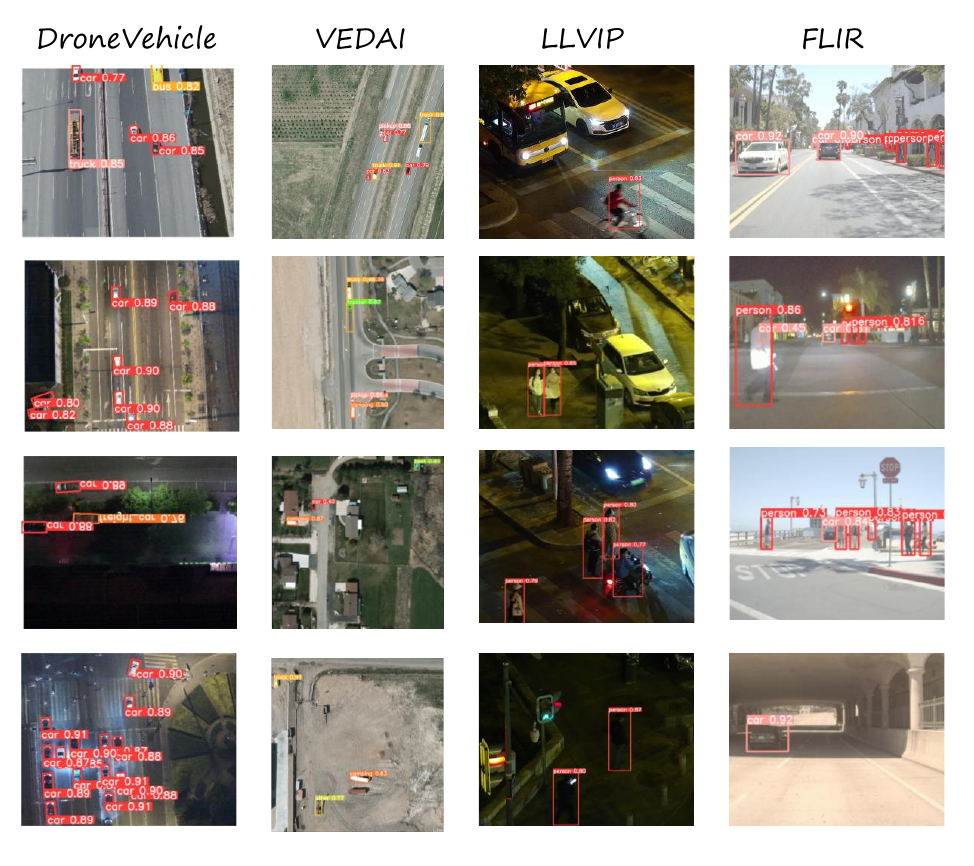}
  \caption{Representative detection results on two aerial RS datasets and two natural scene datasets.}
  \label{VIS}
  % \Description{A woman and a girl in white dresses sit in an open car.}
\end{figure}

The VEDAI dataset presents critical challenges due to fine-grained small objects in complex scenarios. As shown in Table~\ref{tab:vehda}, SuppreSensing achieves the highest mAP@0.5 of 0.840, outperforming the competitive method ADMPF (0.816) by an absolute improvement of 0.024. The improvement is evident for extremely small objects, where our method attains 0.933 AP for trucks (vs. 0.744 for ADMPF) and 0.900 AP for camping cars (vs. 0.847 for ADMPF). Existing strong-alignment fusion methods often introduce excessive semantic consistency, potentially suppressing subtle thermal responses and structural details that are crucial for ultra-small objects. Our modality-specific attribute augmentation strategy alleviates this problem by explicitly modeling bidirectional discrepancy patterns. Together with the ECFP module's iterative refinement mechanism, the model can preserve and strengthen modality-unique semantics for tiny objects while reducing interference from redundant background information.

The DroneVehicle dataset presents critical challenges due to illumination degradation. As shown in Table~\ref{tab:drone}, SuppreSensing achieves a state-of-the-art mAP@0.5 of 0.812, outperforming competitive methods such as MGMF (0.803) and NOC-YOLO (0.795). More importantly, its advantage is more pronounced in challenging categories than in the overall average performance. SuppreSensing achieves an AP50 of 0.823 for trucks, yielding an absolute gain of 12.2\% over MGMF (0.701). This result is consistent with the intended effect of the EMFR module. By casting consensus extraction as an adaptive selection process, EMFR suppresses modality-specific noise, such as optical overexposure and thermal blooming, which often interfere with the representation of large textured objects such as trucks, preserving more reliable multimodal features.

\subsection{Evaluation Across Different Domains}

To further evaluate the generalization ability of SuppreSensing beyond aerial remote sensing, we conduct experiments on two ground-level benchmarks, i.e., LLVIP for extreme low-light surveillance and FLIR for day-night autonomous driving. Compared with aerial imagery, ground-level scenes exhibit substantially different characteristics, including ego-centric viewpoints, heavier occlusions, more complex background clutter, and greater variations in modality quality. As shown in Table~\ref{tab:nal}, SuppreSensing achieves state-of-the-art performance on both datasets, reaching 0.977 mAP@0.5 on LLVIP and 0.760 mAP@0.5 on FLIR, outperforming YOLO-Adaptor and BU-LTT by 1.2\% and 2.8\%, respectively.

The strong cross-domain performance stems from our method not relying on dataset-specific spatial alignment priors but instead modeling multimodal complementarity from both global and local perspectives. Specifically, EMFR adaptively recalibrates modality contributions according to the reliability of the input features, which is particularly important in natural scenes where the quality of visible and thermal signals may vary dramatically across illumination conditions. Meanwhile, ECFP further captures fine-grained cross-modal discrepancies and refines modality-specific spatial cues, enabling the model to preserve discriminative target details under cluttered backgrounds and partial ambiguity. As a result, although natural scenes differ markedly from aerial scenes in viewpoint, scale, and scene layout, the proposed framework remains effective because it is designed to adapt to dynamic modality differences rather than depend on fixed scene assumptions.

\subsection{Visualization of Feature Maps}

Fig. \ref{fig:M-AAM} presents feature visualizations under four illumination conditions in DroneVehicle data. The single-modality features are affected by lighting variations and modality-specific corruption, whereas the fused features better suppress background interference and preserve stable detected object responses. These results verify that SuppreSensing can generate more robust multimodal representations under challenging illumination. Fig. \ref{VIS} shows the visual detection results for two aerial RS datasets and two natural scene datasets. 

\begin{table}[!htbp]
  \renewcommand\arraystretch{1.2}
  \centering
  \caption{Ablation study of the module gain on the DroneVehicle dataset. Best results are in bold.}
  \label{tab:dvab}
  % 修改：使用 \columnwidth 自动适应单栏宽度
  \resizebox{\columnwidth}{!}{
  \begin{tabular}{@{}cc ccccc@{}}
    \toprule
    \multicolumn{2}{c}{EMFR} & ECFP & mAP@0.5 & mAP@0.5:0.95 & Params & GFLOPs \\
    \cmidrule{1-2} 
    % 修改：补齐了缺失的 & 符号，保证一共7列
    CFR & MAA & & & & & \\
    \midrule
        &     & & 0.780 & 0.580 & 16.64M & 69.45\\
    \checkmark &  & &0.804 & 0.609 & 18.63M & 132.42 \\
        & \checkmark & &0.806 & 0.611 & 17.32M & 73.78\\
    \checkmark &\checkmark  & & 0.808 & 0.618 & 18.64M & 132.45 \\
        &  &\checkmark& 0.803 & 0.608 & 28.43M & 105.00 \\
    \rowcolor{gray!20} \checkmark & \checkmark &\checkmark& \textbf{0.812} & \textbf{0.621} & 29.74M & 163.71 \\
    \bottomrule
  \end{tabular}
  }
\end{table}

\begin{table}[!htbp]
  \renewcommand\arraystretch{1.2}
  \centering
  \caption{Ablation study of the module gain on the VEDAI dataset. Best results are in bold.}
  \label{tab:ablation1_vedai}
  
  \resizebox{\columnwidth}{!}{
  \begin{tabular}{@{}cc ccccc@{}}
    \toprule
    \multicolumn{2}{c}{EMFR} & ECFP & mAP@0.5 & mAP@0.5:0.95 & Params & GFLOPs \\
    \cmidrule{1-2} 
    CFR & MAA & & & & & \\
    \midrule
        &     & & 0.763 & 0.501 & 17.04M & 23.26 \\
    \checkmark &  & & 0.806 & 0.531 & 17.05M & 23.27 \\
        & \checkmark & & 0.789 & 0.517 & 18.35M & 42.16 \\
    \checkmark &\checkmark  & & 0.831 & 0.544 & 18.36M & 42.17 \\
        &  &\checkmark& 0.806 & 0.524 & 28.08M & 33.33 \\
    \rowcolor{gray!20} \checkmark & \checkmark &\checkmark& \textbf{0.840} & \textbf{0.554} & 29.39M & 52.24 \\
    \bottomrule
  \end{tabular}
  }
\end{table}

\subsection{Ablation Study}

\noindent \textbf{Effectiveness of Core Modules.} Table~\ref{tab:dvab} and Table~\ref{tab:ablation1_vedai} report the module-wise ablation results on the DroneVehicle and VEDAI datasets, respectively. On the DroneVehicle dataset, the baseline model obtains an mAP@0.5 of 0.780. When CFR, MAA, and ECFP are introduced separately, the mAP@0.5 rises to 0.804, 0.806 and 0.803, respectively. For VEDAI, the metric grows from the baseline value of 0.763 to 0.806, 0.789 and 0.806 respectively. These observations verify that adaptive consensus recalibration, modality-specific discrepancy enhancement and iterative feature purification each contribute to better multimodal feature learning.

Combining CFR and MAA into the full EMFR module further improves mAP@0.5 to 0.808 on DroneVehicle and 0.831 on VEDAI, outperforming either component alone. Adding ECFP further increases mAP@0.5 to 0.812 and 0.840, and mAP@0.5:0.95 to 0.621 and 0.554 on DroneVehicle and VEDAI, respectively. The consistent improvements over the EMFR-only configuration demonstrate that ECFP provides complementary feature refinement following multimodal recalibration.

VEDAI contains a large proportion of tiny objects that occupy merely 0.7\% of the image pixels, rendering their discriminative cues highly susceptible to background clutter. EMFR yields a larger performance gain on VEDAI, improving mAP@0.5 by 0.068 over the baseline, compared to a 0.028 improvement on DroneVehicle. This larger gain suggests that adaptive consensus recalibration and modality-aware discrepancy augmentation are beneficial for preserving weak but discriminative cues of tiny targets. Furthermore, adding ECFP increases the VEDAI mAP@0.5 from 0.831 to 0.840, indicating that progressive purification provides complementary refinement by suppressing residual redundancy and reinforcing task-relevant semantics.

\begin{table}[!htbp]
  \renewcommand\arraystretch{1.2}
  \centering
  \tiny
  \caption{Ablation study on the internal feature paths of the ECFP in the DroneVehicle dataset. Best results are in bold.}
  \label{tab:ecfp_ablation}
  \resizebox{\columnwidth}{!}{
  \begin{tabular}{@{}ccccccc@{}}
    \toprule
    \multirow{2}{*}{EMFR} & \multicolumn{5}{c}{ECFP} & \multirow{2}{*}{mAP@0.5}  \\
    \cmidrule{2-6}
           & SLCP & AMM & SPMP & GSP & EPM & \\
    \midrule
    \checkmark &  &\checkmark &\checkmark &\checkmark &\checkmark & 0.807 \\
     \checkmark &\checkmark &  &\checkmark &\checkmark &\checkmark & 0.810 \\
    \checkmark &\checkmark &\checkmark &  &\checkmark &\checkmark & 0.809 \\
    \checkmark &\checkmark &\checkmark &\checkmark && \checkmark & 0.805 \\
    \checkmark & \checkmark & \checkmark & \checkmark & \checkmark &  & 0.803 \\
     \rowcolor{gray!20}\checkmark &\checkmark &\checkmark &\checkmark  &\checkmark & \checkmark & \textbf{0.812} \\
    \bottomrule
  \end{tabular}
  }
\end{table}

\noindent \textbf{Ablation of ECFP Pathways.} Table~\ref{tab:ecfp_ablation} presents a pathway-wise ablation study to evaluate the contribution of the five customized pathways in ECFP. The complete configuration achieves an mAP@0.5 of 0.812. Removing SLCP, AMM, SPMP, GSP, and EPM reduces the performance to 0.807, 0.810, 0.809, 0.805, and 0.803, respectively. Among these pathways, removing EPM causes the largest drop of 0.9 percentage points, highlighting its importance in maintaining feature continuity across successive refinement stages. Removing GSP results in the second-largest decrease of 0.7 percentage points, indicating that the input-independent global prior helps calibrate feature distribution shifts across network stages. Overall, the performance drops caused by removing any individual pathway confirm the complementary contribution of all five pathways to progressive multimodal feature purification.

% \begin{figure*}
%     \centering
%     \includegraphics[width=7in]{samples/Figs/实验.png}%   
%     \caption{The detection effect of different fusion strategies in the M-SC module (blue circle represents missed detection; green circle represents false detection).}
%      \label{fig:M-AAM}
% \end{figure*}

% \enlargethispage{\baselineskip}
\section{Conclusion}
In this paper, we propose SuppreSensing, an expert-guided feature recalibration and discrepancy augmentation method for multimodal object detection in RS imagery. By integrating input-adaptive recalibration, modality-specific discrepancy modeling, and iterative purification, SuppreSensing suppresses modality-specific interference and mitigates destructive semantic homogenization while enhancing discriminative modality-specific cues. Experiments on DroneVehicle and VEDAI show strong detection performance, while cross-domain evaluations on FLIR and LLVIP confirm its robustness and generalization across diverse conditions.

\begin{acks}
This work was supported in part by the National Natural
Science Foundation of China under Grant U25B2002, in part by the Beijing-Tianjin-Hebei Natural Science Foundation Cooperation Project under Grant No. 25JJJJC0049.
\end{acks}

\bibliographystyle{ACM-Reference-Format}
\balance
\bibliography{sample-base}

\end{document}